\pdfoutput=1

\documentclass[11pt]{article}

\usepackage[]{naacl2021}

\usepackage{times}
\usepackage{latexsym}
\usepackage{subcaption}
\usepackage{graphicx}
\usepackage{amsmath}
\usepackage{amssymb}
\usepackage{url}
\usepackage{float}
\usepackage{caption}

\usepackage{todonotes}
\usepackage{enumitem}

\presetkeys{todonotes}{color=green,size=scriptsize}{}
\newcommand{\Sref}[1]{\S\ref{#1}}

\newif\ifcameraready

\usepackage[T1]{fontenc}

\usepackage[utf8]{inputenc}

\usepackage{microtype}

\title{Case Study: Deontological Ethics in NLP}

\author{Shrimai Prabhumoye~\thanks{\hspace{0.5em} authors contributed equally to this work.}\hspace{0.5em}, Brendon Boldt~\footnotemark[1]\hspace{0.5em}, Ruslan Salakhutdinov, Alan W Black \\
  School of Computer Science \\
  Carnegie Mellon University \\
  Pittsburgh, PA, USA \\
  \texttt{\{sprabhum, bboldt, rsalakhu, awb\}@cs.cmu.edu} \\}

\begin{document}
\maketitle
\begin{abstract}
Recent work in natural language processing (NLP) has focused on ethical challenges such as understanding and mitigating bias in data and algorithms; identifying objectionable content like hate speech, stereotypes and offensive language; and building frameworks for better system design and data handling practices.
However, there has been little discussion about the ethical foundations that underlie these efforts.
In this work, we study one ethical theory, namely deontological ethics, from the perspective of NLP.
In particular, we focus on the \emph{generalization principle} and the \emph{respect for autonomy} through informed consent.
We provide four case studies to demonstrate how these principles can be used with NLP systems.
We also recommend directions to avoid the ethical issues in these systems.
\end{abstract}

\section{Introduction}

The 21st century is witnessing a major shift in the way people interact with technology, and natural language processing (NLP) is playing a central role.
A plethora of NLP applications such as
question-answering systems~\cite{Bouziane2015QuestionAS,gillard-etal-2006-question,yang-etal-2018-hotpotqa} used in diverse fields like 
healthcare~\cite{sarrouti-ouatik-el-alaoui-2017-biomedical,zweigenbaum-2009-knowledge},
education~\cite{godea-nielsen-2018-annotating,raamadhurai-etal-2019-curio},
privacy~\cite{ravichander-etal-2019-question,shvartzshanider-etal-2018-recipe};
machine translation systems~\cite{emnlp-2019-deep,barrault-etal-2019-findings,emnlp-2019-asian,ws-amta-2018-technologies},
conversational agents~\cite{sigdial-2020-special,serban2018survey,liu2016not}, 
recommendation systems~\cite{alharthi2019study,greenquist2019gkb} etc.\ are deployed and used by millions of users.
NLP systems have become pervasive in current human lifestyle by performing mundane tasks like setting reminders and alarms to complex tasks like replying to emails, booking tickets and recommending movies/restaurants.
This widespread use calls for an analysis of these systems from an ethical standpoint.

Despite all the advances in efficiency and operations of NLP systems, little literature exists which broadly addresses the ethical challenges of these technologies.
Ethical theories have been studied for millennia and should be leveraged in a principled way to address the questions we are facing in NLP today.
Instead, the topic of ``ethics'' within NLP has come to refer primarily to addressing bias in NLP systems; \citet{blodgett-etal-2020-language} provides a critical survey of how bias is studied in NLP literature. 
The survey finds that research on NLP systems conceptualize bias differently and that the techniques are not well tied with the relevant literature outside of NLP. 
This creates a gap between NLP research and the study of ethics in philosophy which leaves a rich body of knowledge untapped.

Our work bridges this gap by illustrating how a philosophical theory of ethics can be applied to NLP research.
Ethics (or ethical theory), 
is a theoretical and applied branch of philosophy which studies what is good and right, especially as it pertains to how humans \emph{ought} to behave in the most general sense \citep{iep-ethics}.
As NLP research qualifies as a human activity, it is within the purview of ethics.
In particular, we are using a \emph{prescriptive}, rather than \emph{descriptive}, theory of ethics; prescriptive theories define and recommend ethical behavior whereas descriptive theories merely report how people generally conceive of ethical behavior.

We select two ethical principles from the deontological tradition of ethics and focus on how these principles are relevant to research in NLP.
Namely we look at the \emph{generalization principle} and \emph{respect for autonomy} through informed consent~\citep{sep-kant-moral,kleinig2009}.
We select deonotology because it is reasonable, provides clear ethical rules and comports with the legal idea of the \emph{rule of law} in the sense that these ethical rules bind all persons equally, rather than shifting standards to effect a certain outcome.

We find that there are two main ways in which ethical guidelines can be applied in NLP (or to any other area of technology):
\begin{enumerate}
    \item An ethical guideline can aid in deciding \emph{what} topics within a field merit attention; that is, it answers the question ``which tasks have important ethical implications?''.
    \item An ethical guideline can aid in determining \emph{how} to address a problem; that is, it answers the question ``what factors and methods are preferable in ethically solving this problem?''.
\end{enumerate}
We primarily address (1) and briefly touch on (2) by presenting four case studies relevant to NLP.
In each case study we use an ethical principle to identify an area of research that could potentially conflict with it, and suggest NLP directions to mitigate it.
Although we have selected two principles from a deontological perspective, we are not intimating that these principles can address all ethical issues nor that deontological ethics is the only ethical framework in which our rules and case studies could function (\Sref{sec:discussion}).
Instead, we present the following as a starting point for NLP researchers less familiar but interested in applicable ethical theory.

Our primary contributions are:
\begin{itemize}
    \item Providing an overview of two deontological principles along with a discussion on their limitations with a special focus on NLP.
    \item Illustrating four specific case studies of NLP systems which have ethical implications under these principles and providing a direction to alleviate these issues.
\end{itemize}

\section{Related Work}
\label{sec:related_work}
\subsection{Ethics}
\label{sec:ethics_related_work}

While there are a number of categories of prescriptive ethical theories, including deontology~\citep{kant1785}, consequentialism (e.g., utilitarianism)~\citep{bentham}, and virtue ethics~\citep{aristotle-nico}, we are only addressing deontology.
We do not take a stance in this paper as to whether or not there exists an objectively correct ethical theory, but we offer a brief sketch of deontological ethics and our reasons for using it.
Deontology or deontological ethics refers to a family of ethical theories which hold that whether an act is ethically good or bad is determined by its adherence to ethical rules~\citep{sep-ethics-deontological}.
These rules can be agent-focused duties (e.g., duty to care for one's children) or patient-focused rights (e.g., right to life).
Such rules can also be formulated in modal logic, allowing for more precise reasoning over sets of rules~\citep{hooker-modal2018}.

Deontology stands in contrast to another popular framework of ethics: consequentialism.
Consequentialism holds the ultimate consequences of an action to be the deciding factor regardless of the nature of the actions taken to get there.
We can illustrate the difference between them by observing how each of them might condemn something like racially biased hiring in academia.\footnote{Note that we are presenting generic examples of deontological and consequentialist frameworks and that a variety of nuanced theories in each category exist.}
A deontologist might say that this practice is wrong because it violates the human right to equal treatment regardless of race.
A consequentialist on the other hand, would argue that this is wrong because its \emph{effect} is stymieing academic creativity by reducing intellectual diversity.

We ultimately select the deontological framework in this work for the following reasons:

\begin{enumerate}
    \item We find deontology to be convincing in its own right, namely, its ability to delineate robust duties and rights which protect the value of each and every person.
    \item The universally applicable rules\footnotemark{} of deontology provide a good basis for providing recommendations to researchers.
        \footnotetext{While determining rules which apply universally across all cultures is a difficult task, the existence of organizations, such as the United Nations, presuppose the achievability of identifying internationally applicable norms.}
    Since rights and duties (at their core) are not situation dependent, they are tractable to address in NLP applications.
        \footnote{In contrast to (action-based) utilitarianism which mandates evaluating the full consequences of each action.}
    \item The focus on rights and duties which apply to everyone equally fits well with the widespread legal concept of the \emph{rule of law} which states that every person is subject to the same laws.
\end{enumerate}

\subsection{Ethics in NLP}

We appeal to the fact that problems should be analyzed with a systematic framework, and ethical theories provide precisely these frameworks.
Research should not be based on preconceived notions of ethics which can be overly subjective and inconsistent.
To more rigorously determine what is right and wrong, we rely on ethical theories.
\citet{card2020} present an analysis of ethics in machine learning under a consequentialist framework.
This paper is a kindred spirit in that we both seek to make a philosophical theory of ethics concrete within machine learning and NLP, yet the methods of the paper are somewhat orthogonal.
\citet{card2020} provide a comprehensive overview of how the particular nature of consequentialist ethics is relevant to machine learning whereas we intend to provide tangible examples of how deontological ethical principles can identify ethically important areas of research.
\citet{saltz-ml-ethics2019,bender-etal-2020-integrating} advocate for explicitly teaching ethical theory as a part of machine learning and NLP courses; the case studies in this paper would be a logical extension of the material presented in such a course.

NLP research on ethics has primarily focused on two directions: (1) exploring and understanding the impact of NLP on society, and (2) providing algorithmic solutions to ethical challenges.

\citet{hovy-spruit-2016-social} started the conversation about the potential social harms of NLP technology. 
They discussed the concepts of \textit{exclusion, overgeneralization, bias confirmation, topic under- and overexposure}, and \textit{dual use} from the perspective of NLP research.
A lot of work followed this discussion and made contributions towards ethical frameworks and design practices~\cite{leidner-plachouras-2017-ethical,parra-escartin-etal-2017-ethical,prabhumoye2019principled,schnoebelen-2017-goal,schmaltz-2018-utility}, data handling practices~\cite{lewis-etal-2017-integrating,mieskes-2017-quantitative} and specific domains like education~\cite{mayfield-etal-2019-equity,loukina-etal-2019-many}, healthcare~\cite{suster-etal-2017-short,benton-etal-2017-ethical} and conversational agents~\cite{cercas-curry-rieser-2018-metoo,henderson2018ethical}.
Our paper does not focus on a particular domain but calls for attention towards various NLP systems and what ethical issues may arise in them.

Most of the work providing algorithmic solutions has been focused on bias in NLP systems.
\citet{shah-etal-2020-predictive,tatman-2017-gender,larson-2017-gender} aim to study the social impact of bias in NLP systems and propose frameworks to understand it better. 
A large body of work~\cite{bolukbasi2016man,sun-etal-2019-mitigating,zhao-etal-2019-gender,zhao-etal-2017-men,sap-etal-2019-risk,criticalRaceMethod2020,davidson-etal-2019-racial} directs its efforts to mitigate bias in data, representations, and algorithms.
\citet{blodgett-etal-2020-language} provide an extensive survey of this work and point out the weaknesses in the research design. It makes recommendations of grounding work analyzing bias in NLP systems in the relevant literature outside of NLP, understanding why system behaviors can be harmful and to whom, and engaging in a conversation with the communities that are affected by the NLP systems.
Although issues with bias are certainly within the scope of the principles we present, we do not specifically write on bias because it has already received a large amount of attention.

\section{Deontological Ethics}

There is a variety of specific deontological theories which range from having one central, abstract principle~\citep{kant1785} to having a handful of concrete principles~\citep{ross-rg}.
Rather than comprehensively addressing one theory, we select two rules, one abstract and one concrete, which can fit within a variety of deontological theories. 
The \emph{generalization principle} is an abstract, broad-reaching rule which comes from traditional Kantian ethics.
The \emph{respect for autonomy} is concrete and commonly seen in politics and bioethics.


\subsection{Generalization Principle}

The generalization principle has its roots in Immanuel Kant's theory of deontological ethics~\citep{kant1785}.\footnote{It is also referred to as the ``universal law'' formulation of Kant's categorical imperative.}
The generalization principle states the following~\citep{sep-kant-moral}.
\begin{itemize}[label={}]
    \item An action $\mathcal{A}$ taken for reasons $\mathcal{R}$ is ethical if and only if a world where all people perform $\mathcal{A}$ for reasons $\mathcal{R}$ is conceivable.
\end{itemize}
It is clearer when phrased in the negative.
\begin{itemize}[label={}]
    \item An action $\mathcal{A}$ taken for reasons $\mathcal{R}$ is \emph{un}ethical if and only if a world where all people perform $\mathcal{A}$ for reasons $\mathcal{R}$ logically contradicts $\mathcal{R}$.
\end{itemize}
The main utility of the generalization principle is that it can identify unethical actions that may seem acceptable in isolated occurrences but lead to problems when habitually taken by everyone.

For example, let us take making and breaking a legal contract (the action) whenever it is convenient (the reasons); implicit in the reasons for making a contract is that the other person believes we will follow through~\citep{sep-kant-moral}.
If we universalize this and conceive of a world where everyone makes contracts which they have no intent of keeping, no one would believe in the sincerity of a contract.
Hence, no one would make contracts in the first place since they are never adhered to.
This is the sort of contradiction by which the generalization principle condemns an action and the rationale behind it.

Another example is plagiarism of research papers in conference submissions. Let us assume that a top tier conference did not check for plagiarism because they trust in the honesty of the researchers. In this case, a researcher $\mathbf{G}$ decides to take an action $\mathcal{A}$ of plagiarising a paper due to the following set of reasons $\mathcal{R}$: (1) $\mathbf{G}$ believes that they would not get caught because the conference does not use plagiarism detection software, (2) publishing this paper in the said conference would boost $\mathbf{G}$'s profile by adding $100$ citations, and (3) this would increase $\mathbf{G}$'s chances of getting a job. Plagiarism in this case would be ungeneralizable and hence unethical. If all researchers who want to boost their profile were to submit plagiarised papers, then every researcher's profile would be boosted by $100$ citations, and $100$ citations would lose their value. Hence, this would not increase $\mathbf{G}$'s chances of getting a job, contradicting $\mathcal R3$. Thus, $\mathbf{G}$'s reasons for plagiarism are inconsistent with the assumption that everyone with same reasons plagiarises.

\subsection{Respect for Autonomy}


Respect for autonomy generally addresses the right of a person to make decisions which directly pertain to themselves.
One of the primary manifestations of this is the concept of \emph{informed consent}, whereby a person $\mathbf{A}$ proposes to act in some way $\mathbb{X}$ on person $\mathbf{B}$ which would normally infringe on $\mathbf{B}$'s right to self-govern.
Specifically, we use the formulation of informed consent given by~\citet{pugh-autonomy2020} based on~\citet{kleinig2009}:
\begin{enumerate}
    \item $\mathbf{B}$ must be sufficiently informed with regards to the relevant facts concerning $\mathbb{X}$ to understand what $\mathbb{X}$ is (and what consequences are likely to occur as a result of $\mathbb{X}$).
    \item On the basis of this information, $\mathbf{B}$ \emph{herself} makes the decision to allow $\mathbf{A}$ to do $\mathbb{X}$.
\end{enumerate}

Informed consent is an important idea in bioethics where it typically applies to a patient's right to refuse treatment (or certain kinds of treatment) by medical personnel.
In routine medical treatments this informed consent might be implicit, since one would not go to the doctor in the first place if they did not want to be treated at all, but in risky or experimental medical procedures, explaining the risks and benefits and obtaining explicit consent would be mandatory.
In this case, the patient's autonomy specifically refers to opting out of medical procedures, and informed consent is a concrete method by which to respect this autonomy.

A non-medical example of respect for autonomy and informed consent would be hiring an interpreter $\mathbf{A}$ for a language that the user $\mathbf{B}$ does not speak.
Under normal circumstances, $\mathbf{B}$'s autonomy dictates that she and only she can speak for herself.
But if she is trying to communicate in a language she does not speak, she might consent to $\mathbf{A}$ serving as an \textit{ad hoc} representative for what she would like to say.
In a high-stakes situation, there might be a formal contract of how $\mathbf{A}$ is to act, but in informal circumstances, she would \emph{implicitly} trust that $\mathbf{A}$ translates what she says faithfully ($\mathbb X$).
In these informal settings, $\mathbf{A}$ should provide necessary information to $\mathbf{B}$ before deviating from the expected behaviour $\mathbb{X}$ (e.g., if the meaning of a sentence is impossible to translate).
Implicit consent is a double-edged sword: it is necessary to navigate normal social situations, but it can undermine the respect for autonomy in scenarios when (1) the person in question is not explicitly informed and (2) reasonable expectations do not match reality.

\section{Applying Ethics to NLP systems}
\label{sec:cases}


We apply the generalization principle in \Sref{sec:QA} and \Sref{sec:objection} and respect for autonomy in \Sref{sec:MT} and \Sref{sec:dialogue}.

\begin{figure*}[t!]
    \centering
    \begin{subfigure}[t]{0.50\textwidth}
        \centering
        \includegraphics[height=1.8in]{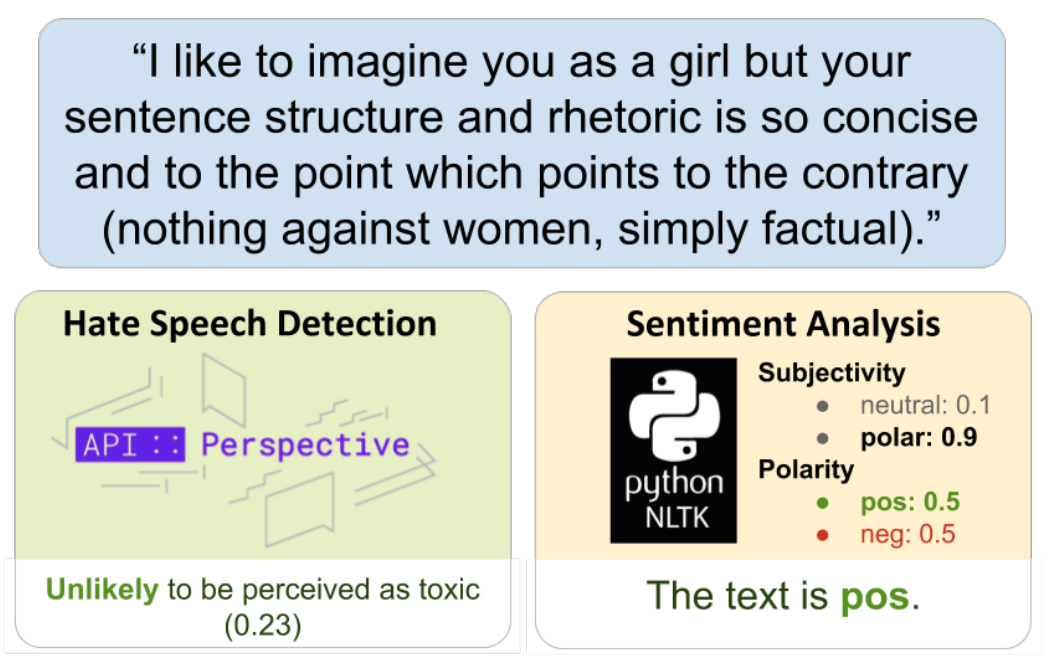}
        \caption{Micro-aggressive comment and its scores by\\ 
        state-of-the-art hate speech detection and \\sentiment analysis tools~\cite{breitfeller-etal-2019-finding}.}
        \label{fig:micro}
    \end{subfigure}%
    \begin{subfigure}[t]{0.50\textwidth}
        \centering
        \includegraphics[height=1.8in]{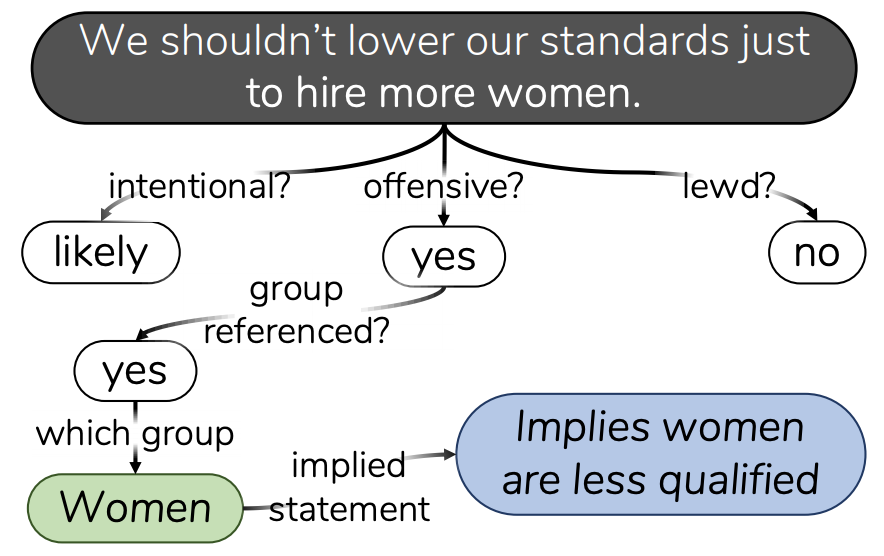}
        \caption{NLP system flagging the micro-aggressive comment as offensive and generating the reasoning for flagging it~\cite{sap-etal-2020-social}.}
        \label{fig:social}
    \end{subfigure}
    \caption{Examples of flagging micro-aggression comments by different NLP systems.}
    \label{fig:microaggression}
\end{figure*}

\subsection{Question-Answering Systems}
\label{sec:QA}
Question-answering (QA) systems have made a huge progress with the recent advances in large pre-trained language models~\cite{devlin-etal-2019-bert,radford2019language,guu2020realm}.
Despite these improvements, it is difficult to know how the model reached its prediction.
In fact, it has been shown that models often obtain high performance by leveraging statistical irregularities rather than language understanding~\cite{poliak-etal-2018-hypothesis,geva-etal-2019-modeling,gururangan-etal-2018-annotation}.
The result is that when a QA system is wrong it is difficult for an end user to determine why it was wrong.
Presumably, the user would not know the answer to the question in the first place, and so it would be difficult to determine even \emph{that} the QA system was wrong.


The act of widely deploying such a QA system is in conflict with the generalization principle.
For example, a QA system $\mathbf{G}$ is unsure of its prediction $\mathcal{A}$ and does not know how it arrived at the answer.
Instead of notifying the user about its inability to reach the prediction, $\mathbf{G}$ decides
to return the prediction $\mathcal{A}$ due to the following reasons $\mathcal{R}$: (1) $\mathbf{G}$ believes that the user does not know the answer and hence (2) $\mathbf{G}$ believes that the user will trust its answer and not ask for reasons for giving the prediction.
If all QA systems operate like this, users will lose trust in QA systems being able to answer their questions reliably and no longer use them.
This contradicts assumption $\mathcal R2$, violating the generalization principle.
This issue goes deeper than a matter of the (in)accuracy of the answer; explainability is still important for a near-perfect QA system.
First, the source of an answer could be fallible (even if the content was interpreted correctly), in which case it is important to be able to point which sources were used.
Second, answers can often be ambiguous, so a user might naturally ask for clarification to be sure of what the answer means.
Finally, it is natural for humans to build trust when working with a system, and explainability is an important step in this process.

Attention weights have been widely used for explaining QA predictions.
Attention weights learnt by neural models denote the words or phrases in a sentence that the model focuses on. 
Hence, words or phrases with high attention weights are considered as explanations to the QA predictions.
But these weights do not reliably correlate with model predictions, making them unsuitable for explainability~\cite{pruthi-etal-2020-learning,serrano-smith-2019-attention,jain2019attention}.
Recently, generating natural language explanations~\cite{rajani2019explain,latcinnik2020explaining} for predictions has gained traction. 
These methods train a language generation model to generate explanations for the QA predictions.
Using a black-box model for text generation, though, pushes the same problem further down the line.
Part of the issue with both of the aforementioned methods is that the ``reasoning'' for the answer is determined \emph{after} the answer has been generated (i.e., reasoning should inform the answer, not vice-versa).

\paragraph{The way forward:}
A method which reaches the prediction through reasoning would be more in line with the generalization principle.
For example, reaching the prediction through traversal of a knowledge graph.
This has been used in scenarios where a knowledge base exists~\cite{Han2020TwoPhaseHB,jansen-etal-2018-worldtree} for a QA system as well as in dynamic graph generation to reach the prediction~\cite{liu2020interpretable,rajagopal2020if,bosselut2019dynamic}.
In these methods, the reasoning is part of the process to generate the final answer, which is more suitable in failing gracefully and building user trust.



\subsection{Detecting Objectionable Content}
\label{sec:objection}

Social media platforms have made the world smaller.
At the same time, the world has seen a surge in hate-speech, offensive language, stereotype and bias on online platforms.
These online platforms have traffic in the millions of textual comments, posts, blogs, etc.\ every day.
Identifying such objectionable content by reading each item is intractable.
Hence, building an NLP system which can read textual data and flag potential objectionable content is necessary.
These systems can reduce the burden on humans by reducing the number of posts that need to be seen by human eyes.


The pivotal role NLP systems play in flagging such content makes the ethical considerations important. 
Fig.~\ref{fig:micro} shows a microaggressive comment and its scores by a state-of-the-art (1) hate speech detection system and (2) sentiment analysis system.
Since these systems rely on surface level words or phrases to detect such (overt) comments, they tend to miss subtle (covert) objectionable content~\cite{breitfeller-etal-2019-finding}. 
If such NLP systems are used universally, then the users of hate speech will discover ways to phrase the same meaning with different words (as illustrated above).
Thus, the NLP content flagging system will not be able to detect objectionable content, and there will be no point in deploying it.
This contradiction suggests that NLP systems must not make their predictions based only on superficial language features but instead seek to understand the intent and consequences of the text presented to them.
Hence, they should generate reasons for flagging posts to facilitate the decision making of the human judges and also to provide evidence about the accuracy of their predictions.


\paragraph{The way forward:} An example of objectionable content is microaggression (Fig.~\ref{fig:microaggression}).
According to Merriam-Webster, microaggression is defined as a ``comment or action that subtly and often unconsciously expresses a prejudiced attitude toward a member of a marginalized group (e.g. racial minority).'' 
Microaggressions are linguistically subtle which makes them difficult to analyze and quantify automatically. 
Understanding and explaining why an arguably innocuous statement is potentially prejudiced requires reasoning about conversational and commonsense implications with respect to the underlying intent, offensiveness, and power differentials between different social groups. 
\citet{breitfeller-etal-2019-finding} provide a new typology to better understand the nature of microaggressions and their impact on different social groups.
Fig.~\ref{fig:social} presents such a comment and how we would like the NLP systems to annotate such content. 
\citet{sap-etal-2020-social} perform the task of generating the consequences and implications of comments which is a step towards judging content based on its meaning and not simply which words it happens to use.
Although such an aim does not automatically solve the problem, attempting to uncover the deeper meaning does not result in an inconsistency or violation of the generalization principle.

\subsection{Machine Translation Systems}
\label{sec:MT}
Machine Translation (MT) systems have reduced language barriers in this era of globalization.
Neural machine translation systems especially have made huge progress and are being deployed by large companies to interact with humans.
But facilitating human-to-human interaction requires more than just simple text-to-text translation, it requires the system to \emph{interpret} the meaning of the language.
This requires a greater sensitivity to style, intent, and context on the part of MT systems.

When an MT system acts as an interpreter for a user, it is essentially speaking for the user when conveying the translated message.
Speaking for one's self is within one's sphere of autonomy, but 
by using the MT system the user has implicitly consented to it representing the user.
That being said, the operating assumption for most users is that the MT system will simply translate the source language into the target language without changing the meaning.
Yet on occasion, differences or ambiguities between languages require either contextual knowledge or further clarification on what is being said.
If the MT system encounters such ambiguities, the user must be \emph{informed} of such occurrences so that she can \emph{consent} to the message which the system ultimately conveys.
Moreover, the user must also be \emph{informed} of the failure cases in the MT system rather than it producing an entirely incorrect translation.

For example, when translating from English to Japanese, there is a mismatch in the granularity of titles or honorifics used to address people.
In English, ``Ms.'' and ``Mr.'' is an appropriate way to address a schoolteacher who does not hold a doctorate.
On the other hand, in Japanese it would be disrespectful to use the more common ``-san'' honorific (the rough equivalent of ``Ms.'' or ``Mr.'') in place of ``-sensei'' which refers specifically to teachers or mentors and shows them a special level of respect.
If the MT system cannot reasonably infer how to resolve the ambiguity in such situations, the English speaker should be \emph{informed} about it.
The English speaker must be notified that such an ambiguity needs to be resolved because there is a risk of offending the Japanese speaker otherwise.

In general, there is a trade-off in translation between literality and fluency in certain situations like the translation of idioms.
Idioms are especially problematic when considering autonomy because there are multiple strategies to translating them which are not only difficult in and of themselves to execute, but deciding which one to use requires the interpreter (i.e., MT system) to understand the intent of the user.
\citet[Ch.~3]{baker1992} identifies five different methods for translating idioms:
\begin{enumerate}
    \item Using an idiom of similar meaning and form; directly translating the idiom achieves the same effect
    \item Using an idiom of similar meaning but dissimilar form; swapping out an equivalent idiom with a different literal meaning
    \item Translation by paraphrase; simply explaining the idiom plainly
    \item Translation by omission
    \item Translation by compensation; for example, omitting idioms in certain locations and adding them in elsewhere to maintain the same overall tone
\end{enumerate}
For example, in casual conversation, an MT system may prefer strategies 1, 2, and 5 to maintain a friendly tone, but in a high-stake business negotiation, it would be more prudent to play it safe with strategy 3.
An MT system must be sensitive to the user's intent since choosing an inappropriate translation strategy could violate her autonomy.

While para-linguistic conduct may fill the gaps for in person interaction, if the interaction is happening only via the textual modality, then there is minimal room for such conduct.
The users in this case may not be aware of the flaws of the MT system representing them.
A recent study~\cite{heinisch-lusicky-2019-user} shows that $45\%$ of the participants reported that they expect MT output, in professional and private contexts, to be useable immediately without any further editing. 
However, post-study, this expectation was not fulfilled. 
The work further shows that the expectation of the type of errors is also different from the errors in the outputs of the MT system. 
For example: only $6\%$ of the participants expect that the output would be useless, but after reading the output, $28\%$ thought that the output was useless. 
The participants in this study had different levels of experience with MT systems (frequent vs.\ rare users) and used MT systems for different functions (private, professional).

\paragraph{The way forward:} 
\citet{Mima1997ImprovingPO} drive the early discussion on using information such as context, social role, domain and situation in MT systems.
~\citet{dimarco1990accounting} 
 advocate for style and intent in translation systems.
A study by \citet{hovy-etal-2020-sound} finds that commercial translation systems make users sound older and more male than the original demographics of the users.
Recent work~\cite{niu2019controlling,sennrich-etal-2016-controlling} has given specific focus to controlling formality and politeness in translation systems.
There is also work directed towards personalizing MT systems~\cite{rabinovich-etal-2017-personalized,michel-neubig-2018-extreme,mirkin-etal-2015-motivating,mirkin-meunier-2015-personalized} while preserving author attributes as well as controlling structural information like voice~\cite{yamagishi-etal-2016-controlling}.
This is a step in the right direction, but we argue that to respect autonomy, translation systems should also obtain explicit informed consent from the user when necessary.

Further research is required in the direction of informing the users about the failure cases of the MT system.
For example, in case of ambiguity, textual interfaces can provide multiple suggestions to the addresser along with the implications of using each variant. 
The user can select the option which best fits their goal. 
In speech interfaces, the MT system can ask a follow up question to the addresser of the system in case of ambiguity or it can add cautionary phrases to the addressee informing them about the ambiguity. 
Alternatively, if the system thinks that the input sentence is ambiguous and cannot be translated with reasonable confidence then it can say “I am unable to translate the sentence in its current form. Can you please rephrase it?”. 
An example scenario where such clarification might be needed is: while translating from English to Hindi if the sentence refers to one’s ``aunt,'' the MT system should ask a follow up question about maternal vs paternal aunt since they have two different words in Hindi language.



\subsection{Dialogue Systems}
\label{sec:dialogue}

We can find a nuanced application of the autonomy principle in the way that dialogue systems, especially smart toys or virtual assistants like Alexa and Google Home, interact with children.

One expression of a parent's autonomy\footnote{This is technically \emph{heteronomy}, but this examples comports with the spirit of \emph{respect for autonomy}.} is generally in deciding whom their child may interact with.
For example a parent would permit interaction with a teacher but not a random stranger.
In the case of a parent purchasing and using a virtual assistant at home, they are implicitly \emph{consenting} to their children interacting with the assistant, and the issue arises from the fact that they may not be \emph{informed} as to what this interaction entails.
To an adult, a virtual assistant or dialogue-capable toy may seem like just another computer, but a 7-year-old child might view it as ``more capable of feelings and giving answers''---a step in the direction of assigning personhood~\citep{druga-eat2017}.
Furthermore, while humans have had thousands of years to learn about human-human interaction, we have only had a half-century to learn about the effects of human-machine (and thus, child-machine) interaction~\cite{mediaEquation}.
We suggest two key areas which are important for dialogue system researchers: (1) they must answer the question of what unique social role do dialogue systems fulfill---that is, in what respects can they be regarded as human-like vs.\ machine-like, and (2) the dialogue systems must have some way of modeling the social dynamics and cues of the interlocutor to fulfill the social role properly.


\paragraph{The way forward:} 
There is a fair amount of research on the social aspects of human-computer dialogue both in general and specifically with regards to children~\citep{druga-eat2017,shen-robot2015,kahn-social2013}.
Although it is difficult to gain a complete understanding of how dialogue systems affect the development of children, the most salient facts (e.g., children regarding virtual assistants as person-like) should be communicated to parents explicitly as part of parental controls.
We advocate for a ``kids mode'' to be included with these virtual AI assistants which would provide the feature of \emph{parental control} in accordance with respect for autonomy.
This mode would be aware that it is talking to children and respond accordingly.
NLP can also help in selecting content and style appropriate for children in these AI agents.
Additionally, parents can be provided with fine-grained control over the topics, sources and language that would be generated by the agent.
For example, the parent can select for a polite language and topics related to science to support their child's development efforts.
Much research has focused on controlling topics~\cite{kim-etal-2015-towards,jokinen-etal-1998-context-management}, style~\cite{niu-bansal-2018-polite}, content~\cite{zhou-etal-2018-dataset,Zhao2020Low-Resource,dinan2018wizard} and persona~\cite{zhang2018personalizing} of dialogue agents which can be used for this purpose.

\section{Ethical Decision Making with NLP}
\label{sec:nlp_aid}

So far we have discussed how NLP systems can be evaluated using ethical frameworks and how decisions made by such systems can be assisted by these theories.
NLP can also aid in making decisions in accordance with the deontological framework.
Recall that the generalization principle judges the ethical standing of pairs of actions and reasons; these pairs could be extracted with various NLP techniques from textual content.
In the case of flagging objectionable content (\Sref{sec:objection}), extracting the deeper intents and implications corresponds to the reasons for the action of flagging the content. 
Another example is building an automatic institutional dialog act annotator for traffic police conversations~\cite{prabhakaran2018detecting}.
These dialog acts contain the rationales of the two agents in the conversation: the police officer and the civilian stopped for breaking traffic rules.
The decision made by the police officer (the action) can then be judged to be in accordance (or not) with a human-selected set of ethically acceptable action and rationale pairs.
Similarly, for court hearing transcripts, the rationales of the arguments can be extracted and the verdict of the judge can be checked using them~\cite{branting2020scalable,workshop-2019-natural-legal}. 
NLP tools such as commonsense knowledge graph generation~\cite{bosselut-etal-2019-comet,saito-etal-2018-commonsense,malaviya2019commonsense}, semantic role~ labeling~\cite{gildea-jurafsky-2000-automatic}, open domain information extraction~\cite{angeli-manning-2013-philosophers} etc.\ can be used to extract rationales, entities from text and also find relations between them to better understand the underlying intent of the text. 


\section{Discussion}
\label{sec:discussion}


We provide a broad discussion on the limitations of the principles chosen in this work and the issue of meta-ethics. 
Moreover, we emphasize that ethical research is not merely a checklist to be satisfied by abiding to the principles mentioned here.
It requires our persistent attention and open-minded engagement with the problem.

One limitation of this work is in the principles that we choose.\footnote{Kant would argue that the generalization principle can account for all ethical decisions, but we make no such claim.}
For example, the interaction of machine learning and privacy is of huge ethical importance.
While the respect for autonomy may address this issue in part, it would be more productive to utilize a deontological principle to the effect of the \emph{right to privacy} with which such matters can be judged.

Another instance is that in this work, we have not discussed the principle of \emph{interactional fairness}~\cite{bies2015interactional,bies2001interactional} which refers to the quality of interpersonal treatment including respect, dignity, and politeness.
With the increasing amount of interaction between humans and machine, the natural language generation systems can be evaluated with this principle.
Systems which show respect and dignity to users as well as generate polite language can enhance the degree of interactional justice, which can in turn enhance utility (e.g., trust, satisfaction).

Additionally, there are broader limitations in using deontology as our ethical framework.
In scenarios where there are no \textit{a priori} duties or rights, taking a consequentialist approach and optimizing the effects of ethical guidelines could be more felicitous.
For example, the specific rights and duties of autonomous AI systems are not immediately clear.
Thus, determining ethical recommendations based on what leads to the most responsible use of the technology would be clearer than selecting appropriate rights and duties directly.
Furthermore, rule-based formulations of consequentialism make ethical judgments based on rules, where the rules are selected based on the consequences.
Such theories combine some of the benefits of both deontology and consequentialism.

The above difficulties are part of the larger issue of metaethics, that is, the discussion and debate on how to choose among different ethical theories.
Within deontology, there is no one standard set of rules.
And even within the generalization principle, there is considerable leeway to what ``conceivable world'' or ``logically consistent'' mean and how they could be applied to decision making.
While presenting a universally accepted ethical theory is likely impossible, metaethical considerations can still be relevant, especially in light of the application of ethical theories.
As the field of NLP gets more accustomed with theories of ethics, it will be fruitful to compare the strengths and weaknesses of different ethical theories within the context of NLP and machine learning.


\section{Conclusion}
\label{sec:conclusion}
Two principles of deontological ethics---namely the \emph{generalization principle} and \emph{respect for autonomy} via \emph{informed consent}---can be used to decide if an action is ethical.
Despite the limitations of these principles, they can provide useful insights into making NLP systems more ethical.
Through the four case studies discussed in this paper, we demonstrate how these principles can be used to evaluate the decisions made by NLP systems and to identify the missing aspects.
For each of the case studies, we also present potential directions for NLP research to move forward and make the system more ethical.

We further provide a summary on how NLP tools can be used to extract reasons and rationales from textual data which can potentially aid deontological decision making.
Note that we do not advocate deontological ethics as the only framework to consider.
On the contrary, we present this work as the first of its kind to illustrate \emph{why} and \emph{how} ethical frameworks should be used to evaluate NLP systems.
With this work, we hope the readers start thinking in two directions: (1) using different ethical frameworks and applying the principles to NLP systems (like the case studies in \Sref{sec:cases}), and (2) exploring the directions mentioned in the case studies of this paper to improve current NLP systems. 

\section*{Acknowledgements}
We are grateful to the anonymous reviewers for their constructive feedback, and special thanks to Dirk Hovy for valuable discussions on this work.
This work was supported in part by ONR Grant N000141812861 and NSF IIS1763562.
This material is based on research sponsored in part by the Air Force Research Laboratory under agreement number FA8750-19-2-0200 (author BB).
The U.S. Government is authorized to reproduce and distribute reprints for Governmental purposes notwithstanding any copyright notation thereon. 
The views and conclusions contained herein are those of the authors and should not be interpreted as necessarily representing the official policies or endorsements, either expressed or implied, of the Air Force Research Laboratory or the U.S. Government.

\bibliography{anthology,custom}
\bibliographystyle{acl_natbib}

\appendix



\end{document}